\documentclass[pdflatex]{sn-jnl}

\usepackage{subfigure}
\usepackage{printlen}
\usepackage{caption}
\usepackage{lineno}
\usepackage{color}
\usepackage{multirow}
\usepackage{makecell}

\jyear{2023}%

\theoremstyle{thmstyleone}%
\theoremstyle{thmstyletwo}%

\theoremstyle{thmstylethree}%

\begin{document}

\setlength{\parindent}{0pt}


\title[]{Judging a video by its bitstream cover}


\author*[1]{\fnm{Yuxing} \sur{Han}}\email{yuxinghan@sz.tsinghua.edu.cn}
\author[2]{\fnm{Yunan} \sur{Ding}}\email{yunanding@eias.ac.cn}
\author[3]{\fnm{Jiangtao} \sur{Wen}}\email{jtwen@rayshaper.ch}
\author[1]{\fnm{Chen Ye} \sur{Gan}}\email{reachfelixgan@gmail.com}


\affil[1]{\orgdiv{Shenzhen International Graduate School}, \orgname{Tsinghua University}, \orgaddress{\city{Shenzhen},
\state{Guangdong}, \country{China}}}
\affil[2]{\orgdiv{Ningbo Institute of Digital Twin}, \orgname{Eastern Institute of Technology}, \orgaddress{\city{Ningbo},
\state{Zhejiang}, \country{China}}}
\affil[3]{\orgname{Shenzhen Tsinghua Research Institute}, 
\orgaddress{\city{Shenzhen}, \state{Guangdong}, \country{China}}}

\abstract{
Classifying videos into distinct categories, such as Sport and Music Video, is crucial for multimedia understanding and retrieval, especially in an age where an immense volume of video content is constantly being generated. Traditional methods require video decompression to extract pixel-level features like color, texture, and motion, thereby increasing computational and storage demands. Moreover, these methods often suffer from performance degradation in low-quality videos. 

We present a novel approach that examines only the post-compression bitstream of a video to perform classification, eliminating the need for bitstream. We validate our approach using a custom-built data set comprising over 29,000 YouTube video clips, totaling 6,000 hours and spanning 11 distinct categories. Our preliminary evaluations indicate precision, accuracy, and recall rates well over 80\%. The algorithm operates approximately 15,000 times faster than real-time for 30fps videos, outperforming traditional Dynamic Time Warping (DTW) algorithm by six orders of magnitude. 

}



\maketitle
\section{Introduction} \label{Intro}

Video classification is fundamental for multimedia services, enabling functionalities such as content retrieval, recommendation, and optimized coding. Traditional techniques focus on analyzing video features such as color, texture, and motion, while recent approaches incorporate deep learning-based algorithms\cite{simonyan2014two}\cite{donahue2015long}\cite{tran2015learning}\cite{zolfaghari2018eco}.

\

A major drawback of current approaches is their dependency on pixel-domain features, resulting in computational and storage inefficiencies. This issue is exacerbated by the high-volume video uploads to platforms like YouTube and TikTok, where videos are typically compressed. Extracting pixel data from such compressed videos necessitates full decoding, leading to a storage increase ratio of up to 75:1 for a 1080p30 video compressed at 10 Mbps. Even with optimizations like down-sampling spatial and temporal resolutions and specialized low-memory-footprint classification techniques \cite{bhardwaj2019efficient}\cite{kondratyuk2021movinets}\cite{wang2021knowledge}, the classification of the 30,000 hours of video uploaded to YouTube every hour would necessitate a method operating thousands of times faster than real-time, while consuming hundreds of times more storage. Furthermore, these techniques often falter in classifying low-quality videos and raise privacy issues, as decryption is needed. Some videos, due to DRM policies, cannot even be decrypted during transmission.

\

    

\begin{figure}[htbp]
    \centering
    \includegraphics[width=\textwidth]{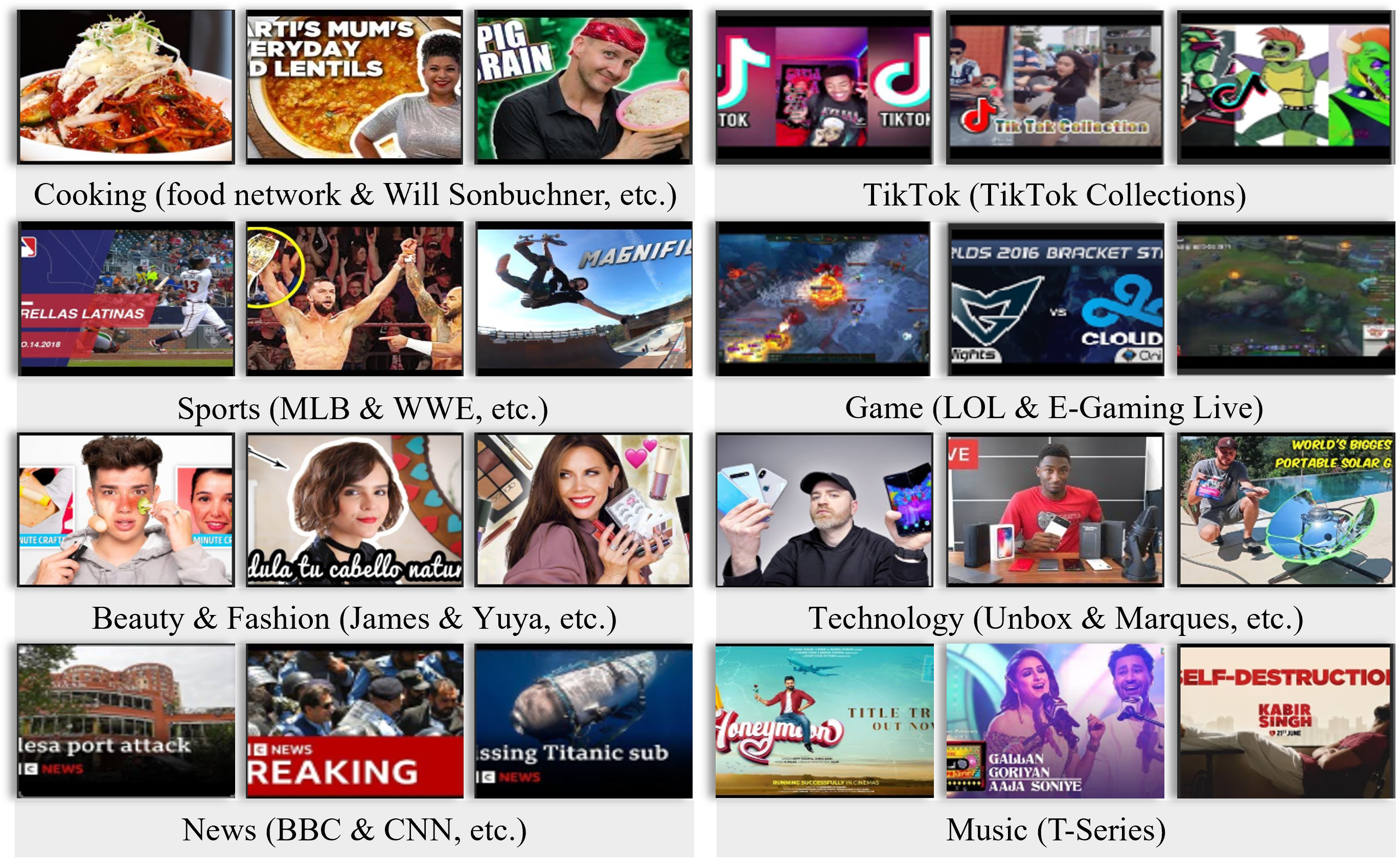}
    \caption{Data Set Thumbnails.}
    \label{fig:thumbnail}
\end{figure}

\begin{figure}[htbp]
    \centering
    \includegraphics[width=\textwidth]{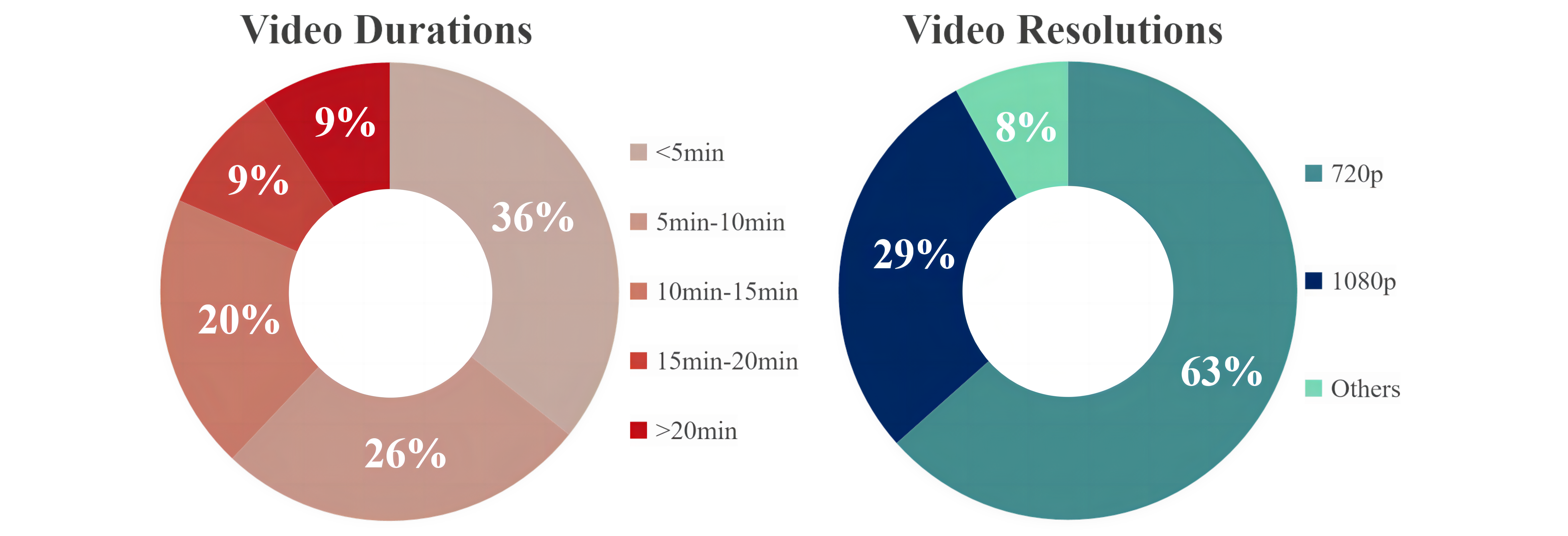}
    \caption{Video Clip Statistics.}
    \label{fig:Video_Clip_Statistics}
\end{figure}

\

In this paper, we introduce a novel direction for video classification that does not rely on pixel domain information. Instead, we use a compressed bitstreams as input for a ResNet-based deep neural network, without the need for bitstream decoding or parsing. This approach leverages the intricate information encapsulated by modern video compression algorithms, particularly advanced spatial and temporal prediction methods found in modern video coding standards such as H.264/AVC \cite{wiegand2003overview}, H.265.HEVC \cite{pastuszak2015algorithm} and H.266/VVC \cite{coding2020iec}. These information include spatial or temporal prediction (“Prediction Mode”), as well as the location of the reference for prediction (e.g. “Motion Vector”), while small number of bits represent prediction error (“residual”). Optimized encoders also incorporate advanced rate control algorithms, which allocates and enforces bitrates across and inside frames. At a very high level, the complexity of individual frames determines the size of the reference frames after compression, the uniformity of the frames determine the size of the predicted frames, whereas the frequency of scene changes is determined by factors such as camera movement rhythms. Utilizing solely this information-rich compressed stream, our methodology significantly reduces computational and storage demands, while ensuring privacy and security.

\

To evaluate the efficacy of our approach, we curated a comprehensive dataset comprising 29,142 video segments across 11 broad YouTube categories, collectively exceeding 6,000 hours in duration. Existing datasets, such as YouTube-8M \cite{abu2016youtube}, 
Activity Net \cite{caba2015activitynet}, UCF101 \cite{soomro2012ucf101}, 
and Net Sports-1M \cite{karpathy2014large}, primarily focus on short-form content, often limiting clips to less than 5 seconds. Our dataset, illustrated in Figure \ref{fig:thumbnail}, is designed to be more encompassing, sourcing through keyword and metadata searches on YouTube. 

\

Through experiments, we observed precision, accuracy, and recall rates consistently exceeding 80\% across numerous test cases. The algorithm demonstrated particular sensitivity to distinct “editing styles”, including factors such as shot selection, camera angles, and movement. Remarkably, even when each video frame was represented by a singular numerical value, the classifier could effectively distinguish between diverse video categories such as NBA games, football matches, and classical or pop concerts. For categories in which editing styles are more diverse, the proposed method was less effective. However, it was still able to identify specific vloggers on YouTube. 

\

Traditionally, we're cautioned against judging a book by its cover. However, much can be inferred about a book's caliber from its cover — the paper quality, meticulous typesetting, and chosen color palette. In a similar vein, our research reveals that a video's compression encoded bitstream is a treasure trove of information, enabling us to appraise a video by its "bitstream cover" with impressive accuracy. This encoded bitstream, while only a fraction of the size of the original video, originates from a sophisticated encoding process and comprises of hundreds, if not thousands, of data points. The sophisticated encoding process, combined with the encoded bitstream's sheer length, makes it a prime target for deep learning to extract insights. 

\

We believe in the potential of exploring the encoded bitstream for video classification offers. This method is especially beneficial for large-scale video analysis and archive digitization, given its capacity to rapidly process vast data sets. Its efficiency also makes it suitable for  real-time applications such as broadcasting and digital marketing. Nonetheless, limitations exist; the approach struggles with videos of similar editing styles, as seen in Gaming videos, where the predominant style is recorded gameplay accompanied by a floating head commentary, as well as differentiating certain finer nuances, such as discerning between NBA clips featuring different basketball stars. However, recognizing its promise, we have open-sourced our model (see Code Availability), inviting research into encoded bitstreams and development of more sophisticated models aimed at video classification without decompression; details of which can be found in the 'Code Availability' section.

\

\section{Results}

\subsection{Data preparation}

We created a large data set consisting of 29,142 video clips, each
containing at least 3,000 frames,
and their corresponding bitstreams compressed using different encoding 
settings or downloaded from YouTube (“entropy coded covers”). 
The clips were selected by searching for key words, and then chosen
based on high play volume or high attention. 
The clips span 11 large and diverse YouTube channels 
\cite{wiki}, including Movie, Entertainment, Knowledge,
Cooking \& Health, Gaming, Technology, Music, Sports, Beauty \& Fashion, 
News, and Education. Detailed video statistics are shown 
in  Fig. \ref{fig:Video_Clip_Statistics}. 

\

\subsection{Hypothesis verification}

\begin{figure}[htbp]
    \centering
    \includegraphics[width=\textwidth]{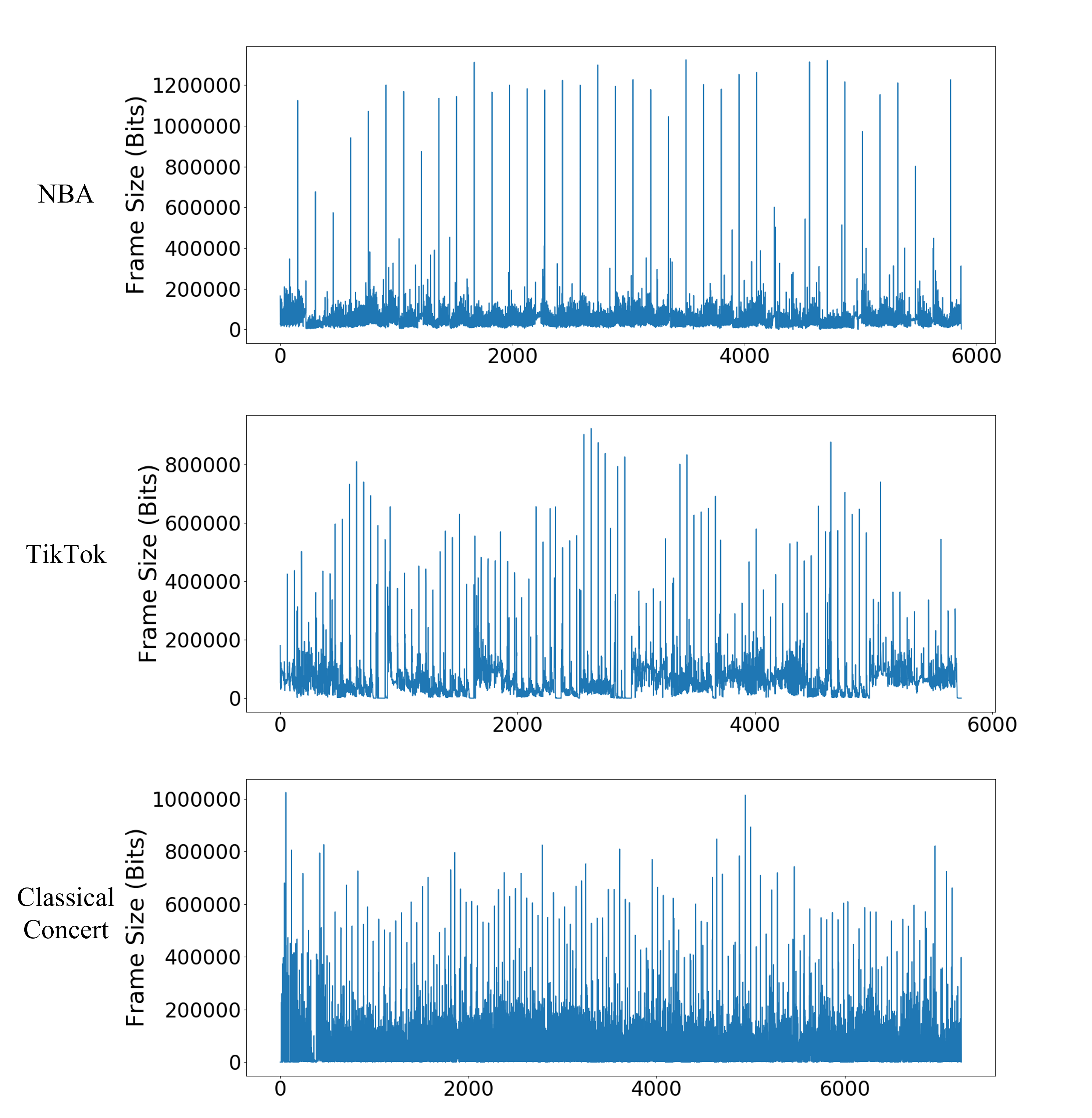}
    \caption{Compressed video size (in bits) as a function of frame number}
    \label{fig:categories_bitrates}
\end{figure}

We propose that videos across various categories exhibit unique style and editing traits, influencing the encoding decisions made by optimized video encoders. These resulting encoded bitstreams may provide revealing high-level statistics, useful for classifying videos with pronounced stylistic differences---such as NBA vs. Football or Pop vs. Classical concerts. This also holds true for content from different social media influencers. Figure \ref{fig:categories_bitrates} substantiates our claim, showing that video clips from various categories display distinct temporal patterns.

\

\begin{figure}[htbp]
    \centering
    \includegraphics[width=\textwidth]{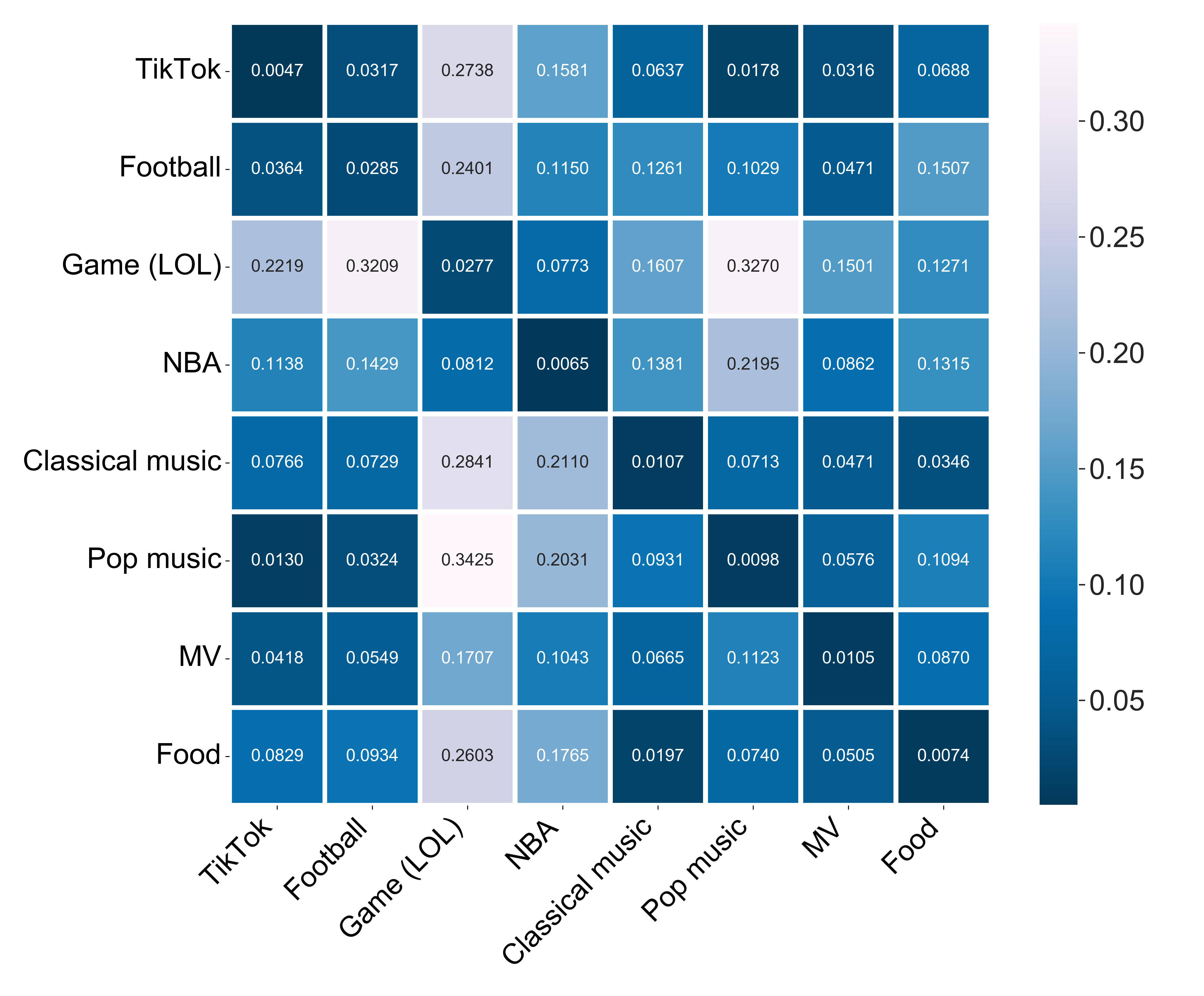}
    \caption{Kullback-Leibler Diverge.}
    \label{fig:Kullback-Leibler Diverge}
\end{figure}

\begin{figure}[htbp]
    \centering
    \includegraphics[width=\textwidth]{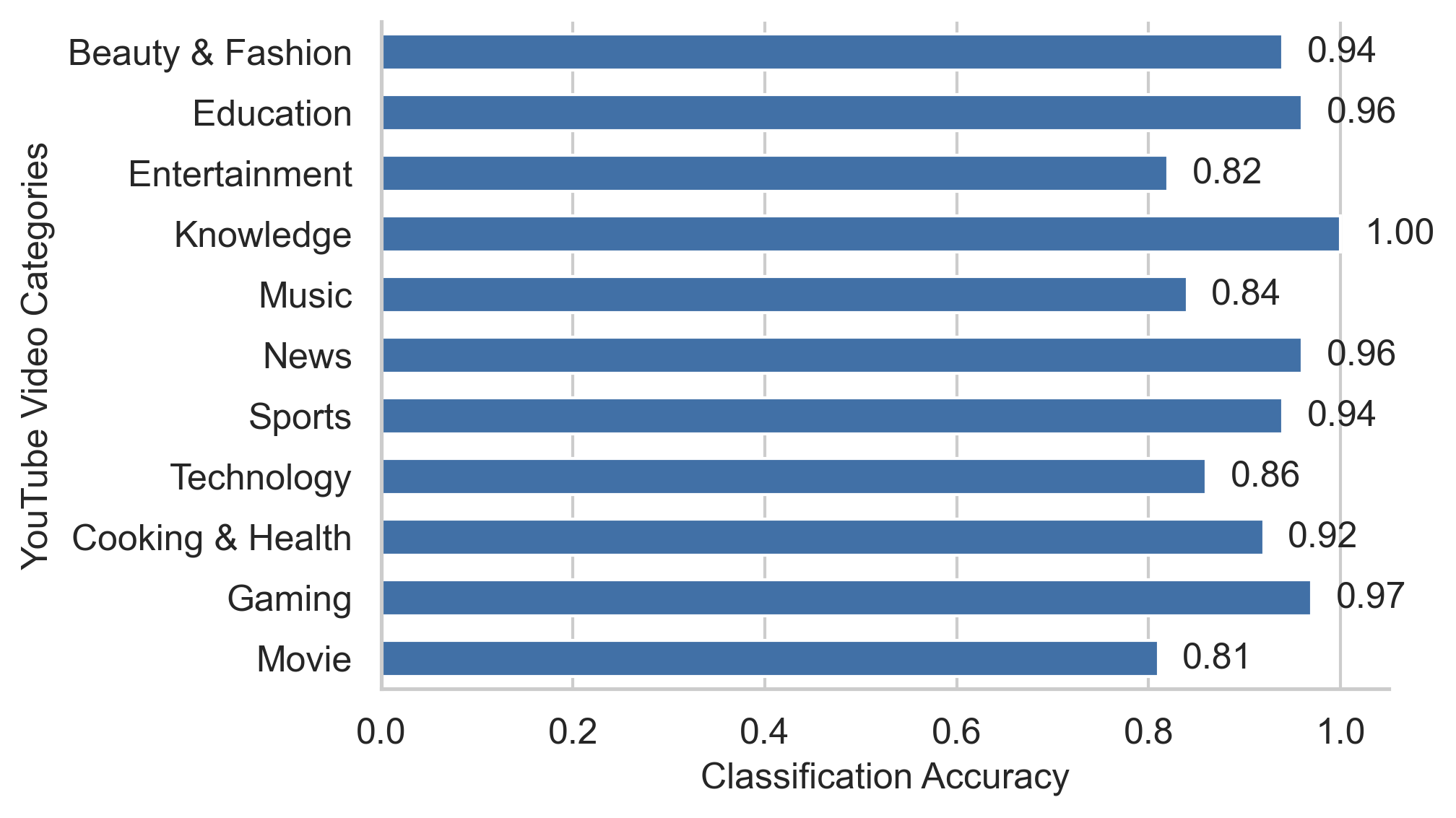}
    \caption{Effectiveness on broad video categories on YouTube}
    \label{fig:11_channel}
\end{figure}

\begin{figure}[htbp]
    \centering
    \includegraphics[width=8cm,height=6cm]{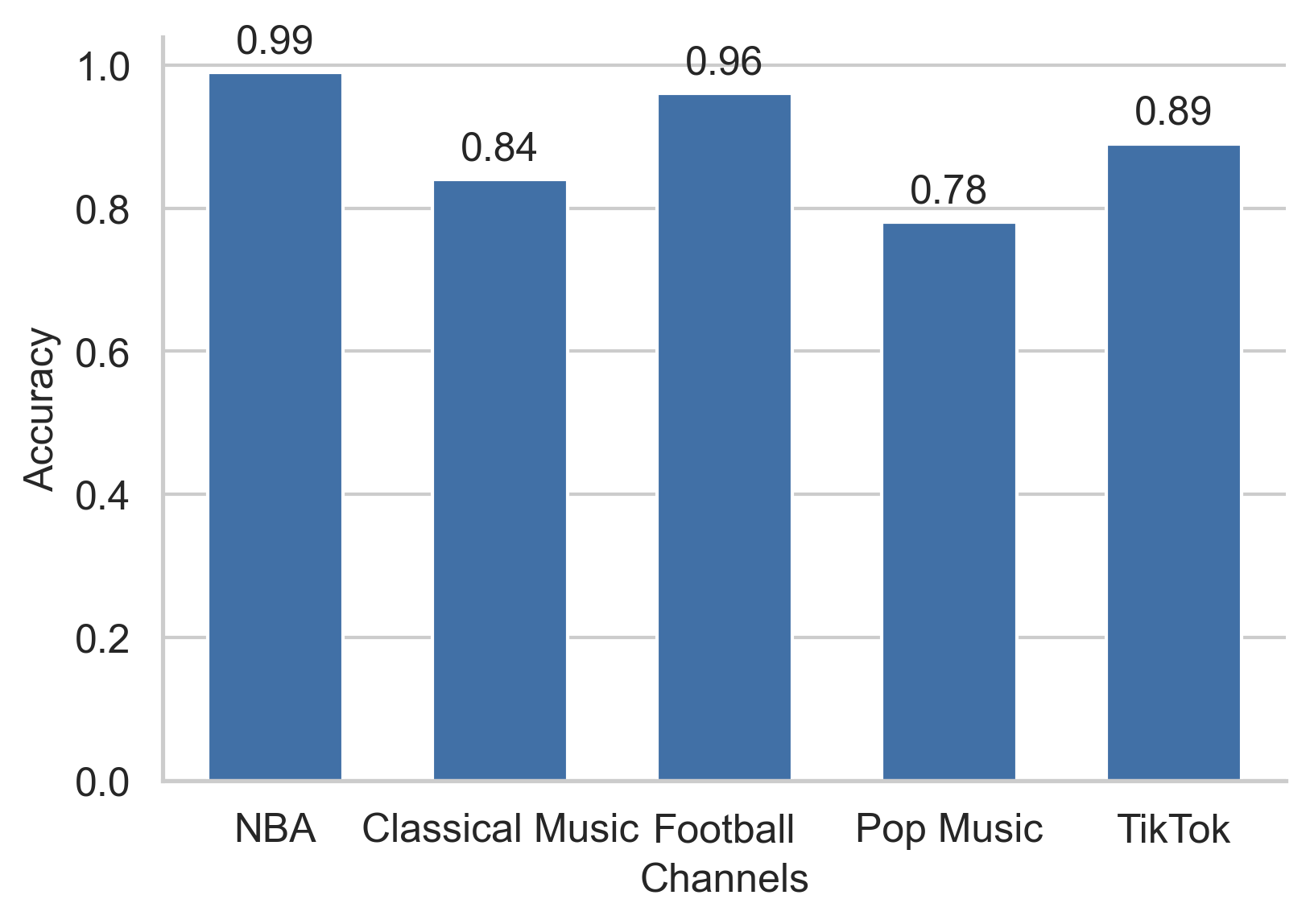}
    \caption{Performance on channels with heavy overlap in content}
    \label{fig:diff_channel}
\end{figure}

\begin{figure}[htbp]
    \centering
    \includegraphics[width=\textwidth]{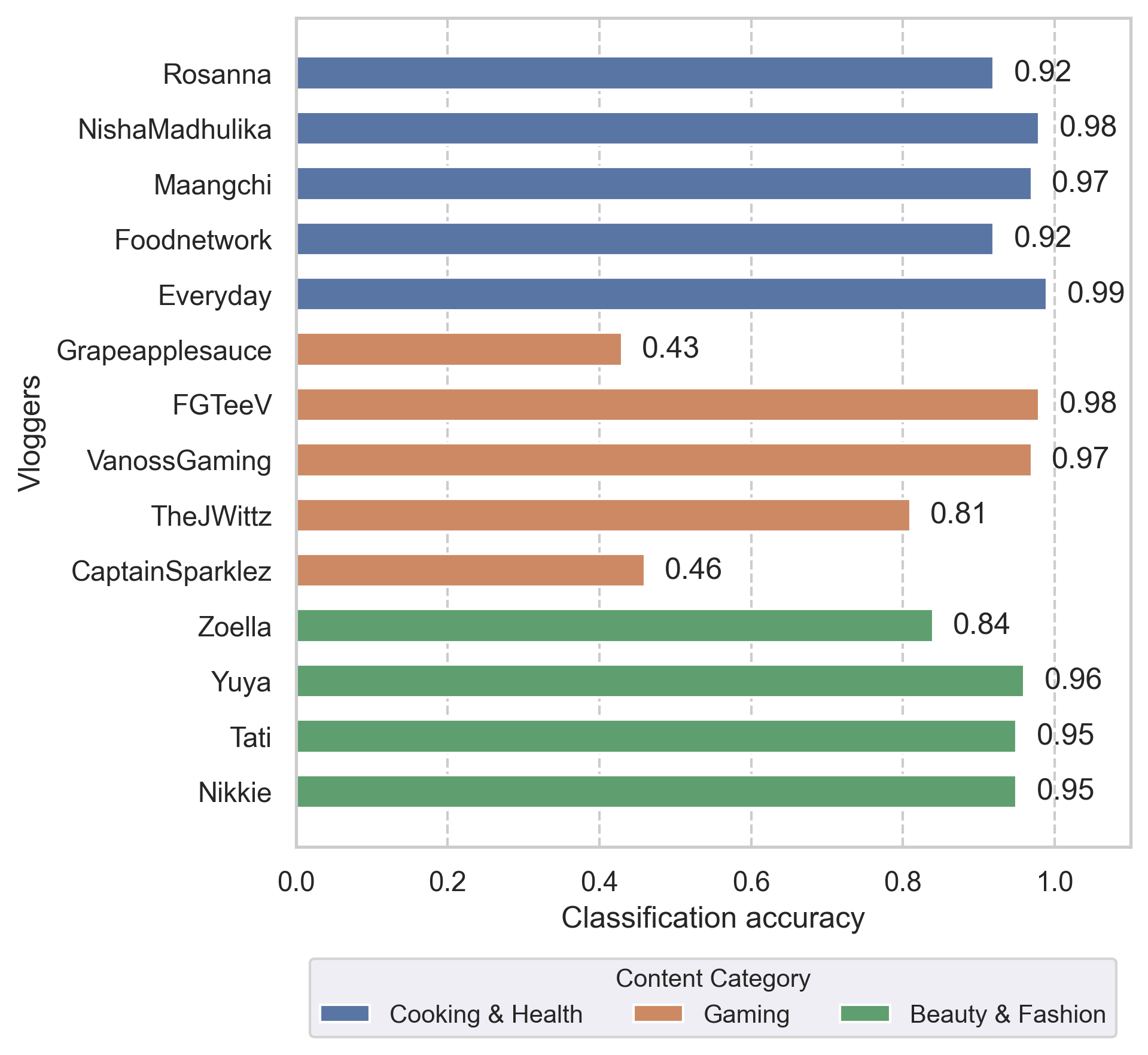}
    \caption{Evaluating ability to discriminate individual vloggers within a channel}
    \label{fig:same_channel}
\end{figure}

We then calculated the 
inter- and intra-class Kullback-Leibler Divergence (KLD) of
the test data set. For clarity and space considerations, we focus on 8 sub-classes, 
namely NBA, TikTok, Football, Classical Concerts, Pop Concerts, 
Gaming (League of Legends), Music Videos, and Culinary 
Exploration. As evident in Fig. \ref{fig:Kullback-Leibler Diverge}, the intra-class KLD—a single, 
first-order metric at the clip-scale level—is typically orders of magnitude
smaller than the KLD between the same class and other classes, despite its limitations in 
capturing long-term and high-order patterns in bitrate time distribution

\

It should also be noted that 
the clips in the categories conceptually overlap, e.g. TikTok might
also contain NBA or Dance clips. We intentionally created such overlaps
to verify the capability of the algorithm to capture the ``main characteristics''
of the content. Because we built the training and testing data sets 
using search and meta-data, we tend to only discover highly popular and viral
clips for each category. For an NBA-themed clip to become viral on
TikTok, it usually follows a certain recognizably TikTok style, 
as a result of the TikTok filters, editing tools and various 
social marketing ``rules''
which makes it distinctively different from a clip of standard NBA broadcast. 
We believe it useful to distinguish an NBA clip as such when it originates from a live broadcast, but to identify that same clip as TikTok content once it has been integrated and disseminated within the TikTok platform. In no cases was the proposed classification algorithm in the loop of test clip class verification and assignment to the training and test data sets.

\subsection{Evaluating classifier performance across diverse data sets}

We employ a ResNet-based classifier\cite{wang2017time} to validate our approach, using the sizes of the encoded frames in bits as the input features. It is worth noting that the reported performance could potentially be enhanced by training a more sophisticated classifier that leverage higher-order input features.

\

To evaluate the robustness and accuracy of our classifier, we carried out tests in three distinct setups: coarse-grained YouTube channel classification, a mixed-category test using clips from channels with similar content, and a fine-grained classification focusing on different vloggers within the same category. 

\

Initially, we evaluate the classifier's performance using high-level YouTube channel classifications, including but not limited to Movie, Entertainment, Knowledge, Cooking \& Health, Gaming, Technology, Music, Sports, Beauty \& Fashion, News, and Education. Notably, the Music category encompasses subcategories like Classical and Pop concerts, T-Series, and music videos, while the Sports category includes NBA, Football, WWE, and MLB among others (see Supplementary Table 1 for more information). Fig. \ref{fig:11_channel} displays that the classification accuracy exceeds 80\% for all categories, and even surpassing 95\% in many instances. The lowest accuracies were recorded in the Movie and Entertainment categories, which naturally have a wide range of content types. 

\

Next, we evaluated the classifier's versatility by examining its performance on TikTok—a platform renowned for its wide array of content. Specifically, we mixed TikTok content with content from various categories prevalent on TikTok, including NBA, Classical Music, Football, and Pop Music (Fig. \ref{fig:same_channel}). The goal was to investigate the classifier's ability to distinguish between, for instance, NBA-related content originating from TikTok and that posted by the NBA itself. We found that the classifier maintained an accuracy rate above 80\% for most categories, with Pop Music being an exception, likely due to its stylistic diversity. Notably, the classifier achieved a respectable 89\% accuracy rate for TikTok content, a finding we attribute to TikTok's widespread use of trending filters. 

\

Finally, we assessed the classifier’s capability for fine-grained distinctions within broad categories like Gaming, Cooking \& Health, and Beauty \& Fashion (Fig. \ref{fig:diff_channel}). The classifier demonstrated over 90\% accuracy in distinguishing vloggers within the Cooking \& Health and Beauty \& Fashion categories. However, it struggled to differentiate between vloggers in the Gaming category. This limitation arises because videos in the Gaming category often combine gameplay screen captures and vlogger commentary, making them inherently similar and thus challenging to classify.

\

Overall, the results confirm the efficacy of ``entropy-encoded covers" as a key feature for video classification. 

\

\subsection{Robustness evaluation across unseen bitrates}

\begin{table*}[]
\resizebox{\textwidth}{!}{
\begin{tabular}{c|ccccccc|c}
\hline\hline
\multirow{2}{*}{\makecell{Performance \\ Metric (\%)}}  
& \multicolumn{6}{c}{\makecell{Test Sets (Bitrate in kbps)}} & & \multirow{2}{*}{\makecell{Training Sets \\ (Bitrate in kbps)}} \\
     & & 500 & 800 & 1000 & 1200 & 1500 & \\ \hline\hline
Precision   &  & {\color[HTML]{FE0000} 85.70} & {\color[HTML]{333333} 75.62} & {\color[HTML]{333333} 68.82} & {\color[HTML]{333333} 64.16} & {\color[HTML]{333333} 54.00} &  &                   \\
Accuracy    &  & {\color[HTML]{FE0000} 85.02} & {\color[HTML]{333333} 66.34} & {\color[HTML]{333333} 50.99} & {\color[HTML]{333333} 39.60} & {\color[HTML]{333333} 25.99} &  & 500               \\
Recall      &  & {\color[HTML]{FE0000} 85.79} & {\color[HTML]{333333} 69.97} & {\color[HTML]{333333} 55.82} & {\color[HTML]{333333} 43.80} & {\color[HTML]{333333} 28.38} &  &                   \\ \hline
Precision   &  & {\color[HTML]{333333} 75.54} & {\color[HTML]{FE0000} 86.87} & {\color[HTML]{333333} 85.35} & {\color[HTML]{333333} 79.39} & {\color[HTML]{333333} 70.15} &  &                   \\
Accuracy    &  & {\color[HTML]{333333} 75.62} & {\color[HTML]{FE0000} 87.62} & {\color[HTML]{333333} 83.54} & {\color[HTML]{333333} 72.40} & {\color[HTML]{333333} 52.35} &  & 800               \\
Recall      &  & {\color[HTML]{333333} 74.93} & {\color[HTML]{FE0000} 88.01} & {\color[HTML]{333333} 84.79} & {\color[HTML]{333333} 75.90} & {\color[HTML]{333333} 56.91} &  &                   \\ \hline
Precision   &  & {\color[HTML]{333333} 71.24} & {\color[HTML]{333333} 86.82} & {\color[HTML]{FE0000} 87.98} & {\color[HTML]{333333} 85.55} & {\color[HTML]{333333} 78.12} &  &                   \\
Accuracy    &  & {\color[HTML]{333333} 70.42} & {\color[HTML]{FE0000} 88.24} & {\color[HTML]{333333} 88.00} & {\color[HTML]{333333} 84.90} & {\color[HTML]{333333} 72.90} &  & 1000              \\
Recall      &  & {\color[HTML]{333333} 70.63} & {\color[HTML]{FE0000} 88.65} & {\color[HTML]{333333} 88.48} & {\color[HTML]{333333} 86.00} & {\color[HTML]{333333} 76.65} &  &                   \\ \hline
Precision   &  & {\color[HTML]{333333} 58.25} & {\color[HTML]{333333} 69.78} & {\color[HTML]{FE0000} 75.19} & {\color[HTML]{333333} 74.86} & {\color[HTML]{333333} 72.25} &  &                   \\
Accuracy    &  & {\color[HTML]{333333} 33.29} & {\color[HTML]{333333} 56.06} & {\color[HTML]{333333} 70.92} & {\color[HTML]{FE0000} 71.53} & {\color[HTML]{333333} 63.00} &  & 1200              \\
Recall      &  & {\color[HTML]{333333} 38.74} & {\color[HTML]{333333} 59.07} & {\color[HTML]{333333} 74.62} & {\color[HTML]{FE0000} 75.56} & {\color[HTML]{333333} 68.84} &  &                   \\ \hline
Precision   &  & {\color[HTML]{333333} 49.41} & {\color[HTML]{333333} 72.67} & {\color[HTML]{333333} 78.55} & {\color[HTML]{333333} 84.76} & {\color[HTML]{FE0000} 85.23} &  &                   \\
Accuracy    &  & {\color[HTML]{333333} 42.45} & {\color[HTML]{333333} 73.14} & {\color[HTML]{333333} 80.45} & {\color[HTML]{FE0000} 85.77} & {\color[HTML]{333333} 85.64} &  & 1500              \\
Recall      &  & {\color[HTML]{333333} 45.61} & {\color[HTML]{333333} 73.68} & {\color[HTML]{333333} 80.56} & {\color[HTML]{333333} 86.30} & {\color[HTML]{FE0000} 86.35} &  &                   \\ \hline\hline
\end{tabular}}
\caption{Classification performance with ABR encoding using B frames at various bitrates}
\label{tab:my-table(ABR)}
\end{table*}

\

We first studied the impact of the mismatch between the bitrate at 
which the ResNet classifier is trained and the bitrate of the 
input clips to the classifiers. Table \ref{tab:my-table(ABR)} shows the performance when the ResNet model was trained 
using 3,000 frames of 1.5Mbps \textit{ 1.5Mbps, 1.2Mbps, 
1.0Mbps, 800kbps, and 500k respectively,} video clips encoded using the average bitrate (ABR) mode, 
for the classification of 3,000 frames of videos encoded also in ABR 
at between 1.5Mbps and 500kbps. 

\

Clearly, classification performance was usually the best when the model
trained at a certain bitrate was used for inputs also at the same bitrate. 
But even for video encoded down to about 50\% of model rate, 
e.g. 800kbps inputs with a model trained at 1.5Mbps,
the classification still performs gracefully.
Only when the bitrate of the input videos to be classified dips 
down to 1/3 of the model training bitrate, does the classification
become a guess.

\subsection{Rate control algorithms and B frames}

To rigorously assess the robustness of our classifier across varying encoding settings, we initially conducted tests using Average Bitrate (ABR) mode (Table \ref{tab:my-table(ABR)}). Subsequently, we explored the Constant Bitrate (CBR) encoding scheme for a subset of 3,000 frames across a bitrate range of 800Kbps to 1.5Mbps to . The performance in both scenarios with and without the use of B-frames is detailed in Tables \ref{tab:my-table(CBR_use_B)} and \ref{tab:my-table(CBR_without_B)}, respectively.

\

Our findings indicate robust classification performance in both conditions. Notably, the incorporation of B-frames yielded even higher accuracy, possibly because B-frames introduce bi-directional correlations in the encoded bitstream. These correlations may better capture the inherent characteristics of the video, thereby enriching the bitstream representation.

\

We further scrutinized the classifier's performance under the Constant Rate Control (CRC) mode, utilizing varying Constant Rate Factors (CRFs) including 0, 18, 23, 28, and 51. The results, with and without B-frames, are cataloged in Table \ref{tab:my-table(CRF_use_B)} and Table \ref{tab:my-table(CRF_without_B)}, respectively. Although the overall classification remained robust, compared with the ABR and CBR mode, the impact of model mismatch was significantly higher. This discrepancy is likely attributable to the perceptual quality model in CRF encoding, which allocates differing weights to motion and texture clarity, thus affecting both encoding choices and the temporal distribution of the bitrate.

\begin{table}[]
\resizebox{\columnwidth}{!}{
\begin{tabular}{c|cllllc|c}
\hline\hline
Performance & \multicolumn{6}{c|}{Test Sets (Bitrate in kbps)}                                                                                & Training Sets     \\ 
Metric (\%)  &  & \multicolumn{1}{c}{800}      & \multicolumn{1}{c}{1000}        & \multicolumn{1}{c}{1200}      & \multicolumn{1}{c}{1500}      &  & (Bitrate in kbps) \\ \hline\hline
Precision   &  & {\color[HTML]{FF0000} 85.11} & 79.66                        & 67.16                        & 58.17                        &  &                   \\ 
Accuracy    &  & {\color[HTML]{FF0000} 85.27} & 74.63                        & 42.20                        & 21.16                        &  & 800               \\ 
Recall      &  & {\color[HTML]{FF0000} 85.72} & 77.27                        & 48.34                        & 29.31                        &  &                   \\ \hline
Precision   &  & 85.16                        & {\color[HTML]{FF0000} 87.93} & 84.80                        & 66.10                        &  &                   \\ 
Accuracy    &  & 86.14                        & {\color[HTML]{FF0000} 87.50} & 80.94                        & 34.78                        &  & 1000                 \\ 
Recall      &  & 86.67                        & {\color[HTML]{FF0000} 88.14} & 82.67                        & 42.65                        &  & \textbf{}         \\ \hline
Precision   &  & 79.19                        & 84.65                        & {\color[HTML]{FF0000} 85.76} & 79.34                        &  &                   \\ 
Accuracy    &  & 80.94                        & 85.40                        & {\color[HTML]{FF0000} 86.26} & 73.76                        &  & 1200               \\ 
Recall      &  & 81.74                        & 86.48                        & {\color[HTML]{FF0000} 87.47} & 76.69                        &  &                   \\ \hline
Precision   &  & 66.07                        & 73.66                        & 81.19                        & {\color[HTML]{FF0000} 84.55} &  &                   \\ 
Accuracy    &  & 61.26                        & 72.15                        & 80.82                        & {\color[HTML]{FF0000} 83.17} &  & 1500               \\ 
Recall      &  & 60.65                        & 71.06                        & 80.70                        & {\color[HTML]{FF0000} 83.24} &  &                   \\ \hline\hline
\end{tabular}}
\caption{Classification performance using CBR
 encoding with B frames at various bitrates}
\label{tab:my-table(CBR_use_B)}
\end{table}

\begin{table}[]
\resizebox{\columnwidth}{!}{
\begin{tabular}{c|cllllc|c}
\hline\hline
Performance & \multicolumn{6}{c|}{Test Sets (Bitrate in kbps)}                                                                                & Training Sets     \\ 
Metric(\%)  &  & \multicolumn{1}{c}{800}      & \multicolumn{1}{c}{1000}     & \multicolumn{1}{c}{1200}     & \multicolumn{1}{c}{1500}     &  & (Bitrate in kbps) \\ \hline\hline
Precision   &  & {\color[HTML]{FF0000} 84.49} & 80.99                        & 70.50                        & 56.61                        &  &                   \\
Accuracy    &  & {\color[HTML]{FF0000} 83.79} & 74.01                        & 46.16                        & 20.67                        &  & 800               \\ 
Recall      &  & {\color[HTML]{FF0000} 85.33} & 76.33                        & 50.64                        & 28.39                        &  &                   \\ \hline
Precision   &  & 78.63                        & {\color[HTML]{FF0000} 83.47} & 80.89                        & 68.80                        &  &                   \\ 
Accuracy    &  & 78.22                        & {\color[HTML]{FF0000} 83.29} & 76.36                        & 42.45                        &  & 1000              \\ 
Recall      &  & 78.47                        & {\color[HTML]{FF0000} 84.44} & 78.62                        & 48.08                        &  & \textbf{}         \\ \hline
Precision   &  & 69.78                        & 77.83                        & {\color[HTML]{FF0000} 82.18} & 79.59                        &  &                   \\ 
Accuracy    &  & 64.11                        & 75.99                        & {\color[HTML]{FF0000} 81.19} & 74.38                        &  & 1200              \\ 
Recall      &  & 64.25                        & 76.32                        & {\color[HTML]{FF0000} 82.81} & 77.94                        &  &                   \\ \hline
Precision   &  & 59.41                        & 72.15                        & 78.54                        & {\color[HTML]{FF0000} 84.11} &  &                   \\ 
Accuracy    &  & 58.04                        & 72.77                        & 79.58                        & {\color[HTML]{FF0000} 82.92} &  & 1500              \\ 
Recall      &  & 55.06                        & 72.01                        & 80.28                        & {\color[HTML]{FF0000} 84.41} &  &                   \\ \hline\hline
\end{tabular}}
\caption{Classification performance using CBR encoding without B frames at various bitrates}
\label{tab:my-table(CBR_without_B)}
\end{table}

\

\begin{table*}[]
\resizebox{\textwidth}{!}{
\begin{tabular}{c|clllllc|c}
\hline\hline
Performance &  & \multicolumn{5}{c}{Test Sets (Constant Rate Factors)}                                                                                                    &  & Training Sets         \\ 
Metric(\%)  &  & \multicolumn{1}{c}{0}        & \multicolumn{1}{c}{18}       & \multicolumn{1}{c}{23}       & \multicolumn{1}{c}{28}       & \multicolumn{1}{c}{51}       &  & Constant Rate Factors \\ \hline\hline
Precision   &  & {\color[HTML]{FF0000} 81.94} & 7.15                         & 21.95                        & 10.01                        & 1.24                         &  &                       \\ 
Accuracy    &  & {\color[HTML]{FF0000} 82.30} & 14.79                        & 13.86                        & 12.87                        & 9.90                         &  & 0                     \\ 
Recall      &  & {\color[HTML]{FF0000} 82.69} & 17.31                        & 15.48                        & 14.99                        & 12.50                        &  &                       \\ \hline
Precision   &  & 16.63                        & {\color[HTML]{FF0000} 84.85} & 77.59                        & 63.32                        & 17.53                        &  &                       \\ 
Accuracy    &  & 11.51                        & {\color[HTML]{FF0000} 85.52} & 78.34                        & 56.44                        & 7.55                         &  & 18                    \\ 
Recall      &  & 11.38                        & {\color[HTML]{FF0000} 86.43} & 81.08                        & 62.07                        & 10.63                        &  &                       \\ \hline
Precision   &  & 12.64                        & 77.59                        & {\color[HTML]{FF0000} 85.13} & 63.32                        & 17.53                        &  &                       \\ 
Accuracy    &  & 17.08                        & 78.34                        & {\color[HTML]{FF0000} 85.77} & 56.44                        & 7.55                         &  & 23                    \\ 
Recall      &  & 13.58                        & 81.08                        & {\color[HTML]{FF0000} 85.88} & 62.07                        & 10.63                        &  &                       \\ \hline
Precision   &  & 12.64                        & 52.06                        & 80.21                        & {\color[HTML]{FF0000} 87.05} & 9.42                         &  &                       \\ 
Accuracy    &  & 17.08                        & 45.79                        & 76.86                        & {\color[HTML]{FF0000} 87.13} & 19.93                        &  & 28                    \\ 
Recall      &  & 13.58                        & 45.79                        & 77.15                        & {\color[HTML]{FF0000} 87.50} & 21.49                        &  &                       \\ \hline
Precision   &  & 0.42                          & 0.42                         & 0.85                         & 27.94                        & {\color[HTML]{FF0000} 86.91} &  &                       \\ 
Accuracy    &  & 3.34                         & 3.34                         & 3.47                         & 5.32                         & {\color[HTML]{FF0000} 86.14} &  & 51                    \\
Recall      &  & 12.50                         & 12.50                        & 12.59                        & 14.27                        & {\color[HTML]{FF0000} 86.82} &  &                       \\ \hline\hline
\end{tabular}}
\caption{Classification performance using different
Constant Rate Factors with B frames}
\label{tab:my-table(CRF_use_B)}
\end{table*}

\begin{table*}[]
\resizebox{\textwidth}{!}{
\begin{tabular}{c|clllllc|c}
\hline\hline
Performance &  & \multicolumn{5}{c}{Test Sets (Constant Rate Factors)}                                                                                                    &  & Training Sets         \\ 
Metric(\%)  &  & \multicolumn{1}{c}{0}        & \multicolumn{1}{c}{18}       & \multicolumn{1}{c}{23}       & \multicolumn{1}{c}{28}       & \multicolumn{1}{c}{51}       &  & Constant Rate Factors \\ \hline\hline
Precision   &  & {\color[HTML]{FF0000} 82.59} & 8.18                         & 6.01                         & 7.50                         & 1.24                         &  &                       \\ 
Accuracy    &  & {\color[HTML]{FF0000} 82.67} & 12.75                        & 10.52                        & 9.90                         & 9.90                         &  & 0                     \\ 
Recall      &  & {\color[HTML]{FF0000} 83.54} & 15.14                        & 13.01                        & 12.42                        & 12.50                        &  &                       \\ \hline
Precision   &  & 6.28                         & {\color[HTML]{FF0000} 81.94} & 77.34                        & 64.87                        & 2.51                         &  &                       \\ 
Accuracy    &  & 17.08                        & {\color[HTML]{FF0000} 82.05} & 74.63                        & 52.48                        & 10.89                        &  & 18                    \\ 
Recall      &  & 14.02                        & {\color[HTML]{FF0000} 83.64} & 77.88                        & 58.58                        & 13.25                        &  &                       \\ \hline
Precision   &  & 3.99                         & 76.41                        & {\color[HTML]{FF0000} 82.30} & 75.76                        & 6.47                         &  &                       \\ 
Accuracy    &  & 17.33                        & 72.77                        & {\color[HTML]{FF0000} 82.30} & 74.38                        & 13.00                        &  & 23                    \\
Recall      &  & 13.15                        & 73.82                        & {\color[HTML]{FF0000} 83.24} & 77.42                        & 15.33                        &  &                       \\ \hline
Precision   &  & 8.53                         & 54.65                        & 78.07                        & {\color[HTML]{FF0000} 84.70} & 10.47                        &  &                       \\ 
Accuracy    &  & 17.20                        & 42.70                        & 73.51                        & {\color[HTML]{FF0000} 83.42} & 16.21                        &  & 28                    \\ 
Recall      &  & 12.76                        & 44.71                        & 75.00                        & {\color[HTML]{FF0000} 84.41} & 17.89                        &  &                       \\ \hline
Precision   &  & 3.61                         & 5.40                         & 19.49                        & 29.24                        & {\color[HTML]{FF0000} 82.48} &  &                       \\ 
Accuracy    &  & 13.12                        & 14.98                        & 20.54                        & 24.63                        & {\color[HTML]{FF0000} 82.67} &  & 51                    \\ 
Recall      &  & 12.99                        & 14.84                        & 18.98                        & 22.55                        & {\color[HTML]{FF0000} 83.58} &  &                       \\ \hline\hline
\end{tabular}}
\caption{Classification performance using different
Constant Rate Factors without B frames }
\label{tab:my-table(CRF_without_B)}
\end{table*}

\subsection{Impact of input size}

\begin{figure}[htbp]
    \centering
    \includegraphics[width=\textwidth]{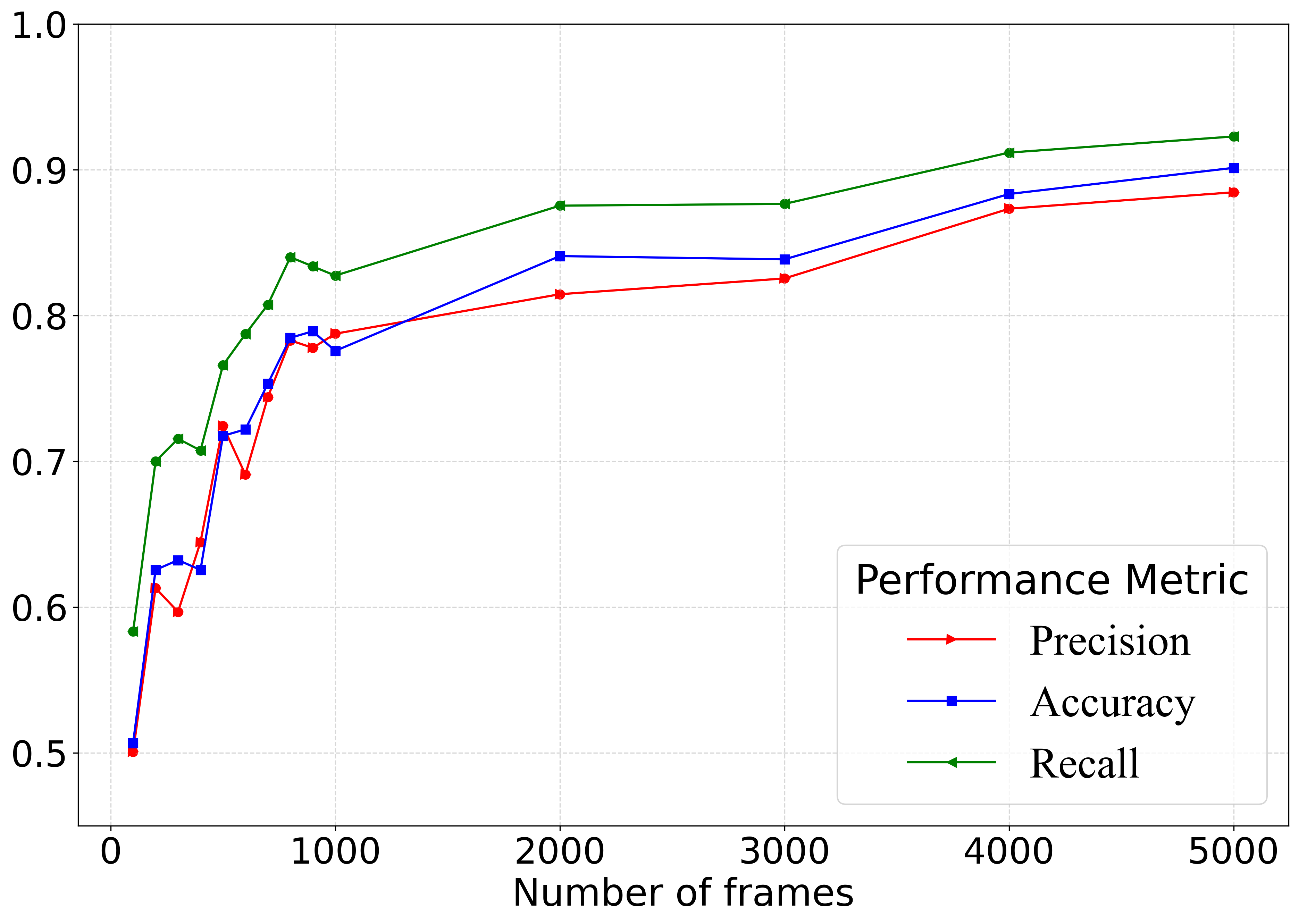}
    \caption{Classification performance as a function of input size.}
    \label{fig:FramePerm}
\end{figure}

\

To assess the classifier's performance as influenced by the number $N$ of frames utilized for model training and classification, we experimented with a range of $N$ values: 120, 240, 360, 480, 600, 720, 840, 960, 1200, 2400, 3600, and 4800 frames. The corresponding classification outcomes for the Average Bitrate (ABR) scenario are depicted in Fig. \ref{fig:FramePerm}. The results indicate that smaller $N$ values may compromise classification performance, likely due to the influence of localized content variations. Conversely, as $N$ increases, classification efficacy enhances and reaches a stable state.

\subsection{Classification speed}
Our ResNet-based video classifier, although not specifically optimized for speed. Even so, on a server equipped with a single Nvidia A100 GPU, it processed 1,818 test videos—each containing 3,000 frames—in less than 13 seconds. This corresponds to a real-time factor of approximately 15,000 for 30fps videos, within striking distance of the 30,000-to-1 ratio of total video duration uploaded to YouTube per unit time. 

\

In contrast, we evaluate the conventional non-deep-learning-based Dynamic Time Warping (DTW) algorithm \cite{berndt1994using}, commonly applied in Time Series Classification (TSC) problems. Utilizing the same dataset and categories but restricting the videos to short 30-frame clips, the DTW algorithm required an extensive 27 hours for classification. Given DTW's computational complexity of $O(N^2)$, where $N$ is the time series length, the algorithm is approximately $7.5 \times 10^7$ times slower than the ResNet classifier. Moreover, the DTW-based accuracy rate is less than 0.35. Although longer time series might slightly improve DTW's performance, the algorithm's inefficiency makes it an impractical alternative for long-sequence time-series forecasting (LSTF) in video classification tasks based on a sequence of encoded frame.

\begin{figure*}[htbp]
    \centering
    \includegraphics[width=\textwidth]{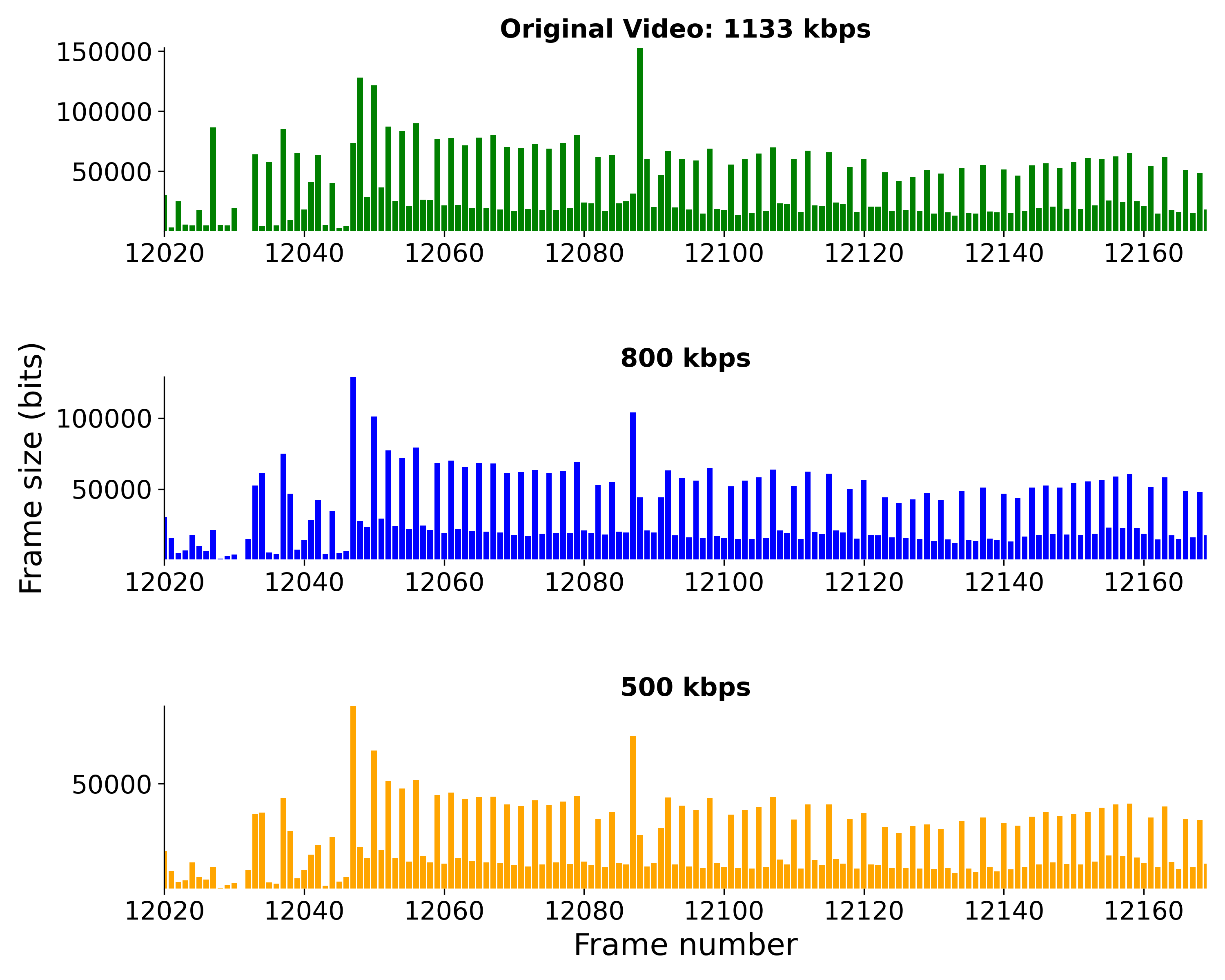}
    \caption{Bitrate variation for the same clip transcoded to different
    bitrates.}
    \label{fig:DiffRate}
\end{figure*}

\section{Discussion}\label{Conclusion}
In this work, we introduce an innovative approach for video classification that leverages compressed bitstream representations of videos, eliminating the need for pixel-level decoding. This approach offers multiple advantages: it minimizes storage requirements, reduces computational overhead, enhances data privacy, and is resilient to performance degradation resulting from poor video quality. 

\

We demonstrate robust classification capabilities across both coarse and fine categories, including those that overlap. Furthermore, it operates with a speed of approximately 15,000 times real-time for videos at 30fps.

\

In our experiments, we employed a straightforward classifier that exclusively utilizes the time series of compressed video frame sizes as input. This information is readily accessible through various means—such as byte-aligned headers or Network Abstraction Layer (NAL) packets—without the need for decoding. This design not only simplifies computational complexity but also enhances data privacy. Remarkably, NAL packet-based analysis could even function with encrypted content, positioning our technique as a scalable solution for network carriers.

\

Although our classifier is still in its nascent stages with room for optimization, preliminary results are promising. One significant factor affecting classification performance is the number of frames ($N$) used for both training and classification. Our ongoing research includes adaptive methods that increment $N$ progressively to improving classification outcomes. Additionally, we are examining the effects of employing diverse encoders and dynamically varying the number of frames in the input clip for classification. In our tests, we intentionally selected $N$ values that do not align with standard Group of Picture sizes (e.g., 30, 60) in an attempt to approximate the effects of adaptive intra-frame insertion. However, this area requires further investigation. 

\

One limitation lies in the need to retrain the classifier when the number of classes changes. We suggest an adaptive model that could categorize new or unanticipated classes under a generic "Others" category, which could then serve as input for a specialized secondary classifier. We also limited our tests to distinguishing between categories with pronounced editing styles, such as Sports from Music Videos, and did not aim to differentiate, for instance, NBA clips featuring Michael Jordan from those featuring LeBron James. We anticipate challenges in identifying such nuanced distinctions, as those are likely obscured during the encoding process. 

\

It is important to note that our work is an initial step and not the definitive solution in using a video's compression encoded bitstream representation for classification. Nevertheless, we believe our early findings demonstrate the considerable potential of this avenue. We have open sourced our model (see Code Availability), and aim to inspire subsequent research and discussion on developing more advanced classifiers that can surpass our current model without pixel-level decoding.


\section{Methods} \label{sec_method}

\subsection{Dataset and preprocessing}

Distinct video categories display unique patterns related to scene transitions, texture, shot lengths, etc. When encoded by a rate-distortion optimized encoder, these patterns translate into varied bitstream distributions and entropy concerning motion vectors, modes, and bitrate allocations. This effect is even more pronounced for social network videos that use preset filters and special effects.

To test our hypothesis, we compiled a video dataset of 29,142 clips from YouTube, spanning 11 categories: Movies, Entertainment, Cooking \& Health, Gaming, Technology, Music, Sports, Beauty \& Fashion, News, and Education. Each category held a minimum of 400 clips, with each clip containing at least 3,000 frames. The clips varied in spatial resolution, from 298x480 for TikTok videos to 720p and 1080p for others, and were consistent in frame rate (either 30fps or 60fps) within their respective categories. The peak video bitrate recorded was 3Mbps. Fig. \ref{fig:categories_bitrates} illustrates the typical size sequences of compressed frames across categories. 

For our experiment, we transcoded the videos to a 1.5Mbps bitrate using FFmpeg's H.264/AVC encoder, maintaining consistent encoding settings across all clips. As shown in Fig. \ref{fig:DiffRate}, a test clip originally downloaded at 1.13Mbps was transcoded to both 1Mbps and 500kbps. While the encoding setting differed from the original YouTube version, the frame size time series of various bitstreams remained highly correlated. This suggests that 1) post-compression frame sizes, reflecting bitrate variations, could be useful in video classification; and 2) it might be feasible to develop a single model capable of classifying videos across a spectrum of bitrates.

\

The Kullback-Leibler Divergence results (Fig.\ref{fig:Kullback-Leibler Diverge}) supports this idea.

\subsection{Problem formulation}

\

Video classification using the time sequence of the sizes of video frames after 
compression (in encoding order) is a typical
a Time Series Classification (TSC) problem \cite{gamboa2017deep} 
\cite{bagnall2017great}\cite{ruiz2021great}.
We define the sizes of compressed frames in bits in encoding order 
as a time series $X = {[x_1,x_2,x_3,...,x_T]} \in \mathbb{R}^T$, where T represents the length of the time series. Consequently, video classification can be formulated as a mapping between $X$ and a one-hot label vector $Y$.

\

Traditional TSC algorithms often use distance-based metrics with 
a $k$-nearest neighbor classifier ($K$-NN). Such algorithms
include Dynamic Time Warping (DTW) \cite{berndt1994using}, Weighted DTW  
\cite{jeong2011weighted}, Move–split–merge \cite{stefan2012move}, Complexity invariant 
distance \cite{batista2014cid}, Derivative DTW \cite{gorecki2013using}, Derivative 
transform distance \cite{gorecki2014no}, Elastic ensemble \cite{lines2015time} , etc., 
where DTW is often used as a baseline \cite{bagnall2017great}. 
All such methods take the entire time 
series as input and are computationally intensive. Even following the reduced
complexity TSC algorithm from Rodríguez et al. \cite{rodriguez2004support}, 
TSC algorithms are in general, still highly time-consuming. 

\

In recent years, deep learning based TSC has been studied broadly. A 
study by Fawaz et al.\cite{ismail2019deep} compared the TSC performances
of Time Convolutional Neural Network (Time-CNN) \cite{zhao2017convolutional},
Encoder \cite{serra2018towards}, Fully Convolutional Neural Networks (FCNs) 
\cite{wang2017time}, Multi Layer Perceptron (MLP) \cite{wang2017time}, 
Residual Network (ResNet) \cite{wang2017time}, 
Multi-scale Convolutional Neural Network (MCNN)  \cite{cui2016multi}, 
Multi Channel Deep Convolutional Neural Network 
(MCDCNN) \cite{zheng2014time}, Time Le-Net (t-LeNet) 
\cite{le2016data}, and Time Warping Invariant Echo State 
Network (TWIESN) \cite{tanisaro2016time}
using the UCR/UEA archive \cite{c.dataset1} and MTS archive \cite{c.dataset2} 
data sets. It was concluded that deep residual network architecture performs the 
best. Based on the above studies, we used DTW as a baseline for traditional 
TSC algorithms, and ResNet as the representative structure for deep
learning based algorithms.

\subsection{Video classification using temporal bitrate variation}

In this section, we provide details for training the time series classifier.

\begin{figure*}[htbp]
    \centering
    \includegraphics[width=\textwidth]{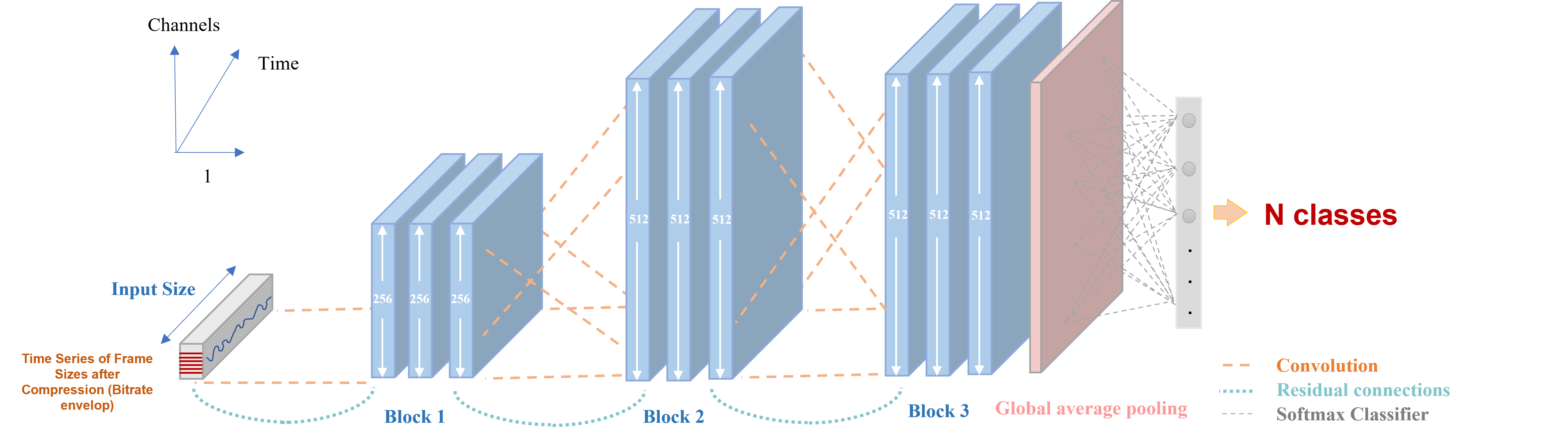}
    \caption{Residual architecture classifier based on \cite{wang2017time}.}
    \label{fig:ResNet}
\end{figure*}

Our primary objective in this study was to explore the potential of bitrate time sequence for video categorization. Consequently, we prioritized this over intricate neural network design and opted for the established ResNet classifier \cite{wang2017time}, depicted in Fig. \ref{fig:ResNet}.

\

\textbf{Model input.} We set the labeled input-target pairs $(X_i,Y_i) \in \mathcal{D}$ as input, where $\mathcal{D} \in \mathbb{R}^{N \times T \times C }$ denotes a set of $N$ compressed video frames represented in bits, each with a fixed length of $T$ and $C$ channels. In contrast to traditional pixel-based approach, our approach treats each frame as a singular value, as opposed to millions of pixel values. This significantly accelerates classification speeds. The input length in our network is determined by the number of video frames used. The channel count is dependent on bitstream configuration. For instance, using simply compressed frame sizes as input yields a single channel. However, incorporating additional bitrate data from various parts of the bitstream, like motion vectors, headers, and texture information, can increase the number of channels.

\

\textbf{Model architecture.} The ResNet based classifier mainly consisted of $3$ residual blocks, which are used to extract features from the input time series. 
To describe the process of a classifier predicting the class of a given input X in this supervised learning task, the equation can be expressed as:
\begin{equation}
    \hat{y} = f_L(\theta_L, x) = f_{L-1}(\theta_{L-1}, f_{L-2}(\theta_{L-2}, \dots, f_1(\theta_1, x + \mathcal{F}(x)))
\end{equation}

where $\hat{y}$ is the predicted output, and $\mathcal{F}(x)$ denotes the shortcut connection in each residual block. The final output $\hat{y}$ of the ResNet model is produced by a global average pooling layer and a softmax classifier. The global average pooling layer generates feature maps of $\hat{y}$  for different video categories in this classification task, then the softmax classifier maps the feature vector to a probability distribution over the output classes. The main characteristic of the network is the residual connection between different convolutional layers, which effectively avoids vanishing gradient in the training process \cite{he2016deep} \cite{wang2017time}. The number of filters for each convolutional blocks are set to 256, 512, 512 respectively.

\

\textbf{Model Training Specifics.} We used fixed parameter initialization and an auto-adjusted learning rate, with the factor at 0.5 and the patience level at 40. To avoid overfitting, an early stop strategy was adopted with the patience set to 80. The adam optimizer was used for all contrast experiments.

\

For this multi-classification task, categorical cross-entropy loss function is used as the supervised training loss, which is written as

\begin{equation} \label{eq:cross-loss} \mathrm{Loss}(\mathbf{y}, \mathbf{\hat{y}}) = -\sum_{i=1}^{C} y_i \log(\hat{y}_i) \end{equation} 

where $y_i$ denotes one-hot encoding for the corresponding video category. $C$ denotes the total number of categories. The optimization goal of Equation \ref{eq:cross-loss} is to maximize the likelihood of predicting true labels given the model parameters.


\backmatter





\section{Data availability}

Source data are provided at the following link: 
\url{https://tinyurl.com/bitstream-video-data}

\section{Code availability}

The code for reproducing the results is available from GitHub at:  \url{https://tinyurl.com/bitstream-video-code}









\bibliographystyle{naturemag}
\bibliography{references}

\begin{thebibliography}{10}
\expandafter\ifx\csname url\endcsname\relax
  \def\url#1{\texttt{#1}}\fi
\expandafter\ifx\csname urlprefix\endcsname\relax\def\urlprefix{URL }\fi
\providecommand{\bibinfo}[2]{#2}
\providecommand{\eprint}[2][]{\url{#2}}

\bibitem{simonyan2014two}
\bibinfo{author}{Simonyan, K.} \& \bibinfo{author}{Zisserman, A.}
\newblock \bibinfo{title}{Two-stream convolutional networks for action
  recognition in videos}.
\newblock In \emph{\bibinfo{booktitle}{28th Conference on Neural Information
  Processing Systems (NIPS)}}, vol.~\bibinfo{volume}{27}
  (\bibinfo{year}{2014}).

\bibitem{donahue2015long}
\bibinfo{author}{Donahue, J.} \emph{et~al.}
\newblock \bibinfo{title}{Long-term recurrent convolutional networks for visual
  recognition and description}.
\newblock In \emph{\bibinfo{booktitle}{IEEE Conference on Computer Vision and
  Pattern Recognition (CVPR)}}, \bibinfo{pages}{2625--2634}
  (\bibinfo{year}{2015}).

\bibitem{tran2015learning}
\bibinfo{author}{Du, T.} \emph{et~al.}
\newblock \bibinfo{title}{Learning spatiotemporal features with 3d
  convolutional networks}.
\newblock In \emph{\bibinfo{booktitle}{IEEE International Conference on
  Computer Vision}}, \bibinfo{pages}{4489--4497} (\bibinfo{year}{2015}).

\bibitem{zolfaghari2018eco}
\bibinfo{author}{Zolfaghari, M.}, \bibinfo{author}{Singh, K.} \&
  \bibinfo{author}{Brox, T.}
\newblock \bibinfo{title}{Eco: Efficient convolutional network for online video
  understanding}.
\newblock In \emph{\bibinfo{booktitle}{Proceedings of the European conference
  on computer vision (ECCV)}}, \bibinfo{pages}{713--730}
  (\bibinfo{year}{2018}).

\bibitem{bhardwaj2019efficient}
\bibinfo{author}{Bhardwaj, S.}, \bibinfo{author}{Srinivasan, M.} \&
  \bibinfo{author}{Khapra, M.~M.}
\newblock \bibinfo{title}{Efficient video classification using fewer frames}.
\newblock In \emph{\bibinfo{booktitle}{Proceedings of the IEEE/CVF Conference
  on Computer Vision and Pattern Recognition (CVPR)}},
  \bibinfo{pages}{354--363} (\bibinfo{year}{2019}).

\bibitem{kondratyuk2021movinets}
\bibinfo{author}{Kondratyuk, D.} \emph{et~al.}
\newblock \bibinfo{title}{Movinets: Mobile video networks for efficient video
  recognition}.
\newblock In \emph{\bibinfo{booktitle}{Proceedings of the IEEE/CVF Conference
  on Computer Vision and Pattern Recognition (CVPR)}},
  \bibinfo{pages}{16020--16030} (\bibinfo{year}{2021}).

\bibitem{wang2021knowledge}
\bibinfo{author}{Wang, L.} \& \bibinfo{author}{Yoon, K.-J.}
\newblock \bibinfo{title}{Knowledge distillation and student-teacher learning
  for visual intelligence: A review and new outlooks}.
\newblock \emph{\bibinfo{journal}{IEEE transactions on pattern analysis and
  machine intelligence}} \textbf{\bibinfo{volume}{44}},
  \bibinfo{pages}{3048--3068} (\bibinfo{year}{2021}).

\bibitem{wiegand2003overview}
\bibinfo{author}{Wiegand, T.}, \bibinfo{author}{Sullivan, G.~J.},
  \bibinfo{author}{Bjontegaard, G.} \& \bibinfo{author}{Luthra, A.}
\newblock \bibinfo{title}{Overview of the h. 264/avc video coding standard}.
\newblock \emph{\bibinfo{journal}{IEEE Transactions on circuits and systems for
  video technology}} \textbf{\bibinfo{volume}{13}}, \bibinfo{pages}{560--576}
  (\bibinfo{year}{2003}).

\bibitem{pastuszak2015algorithm}
\bibinfo{author}{Pastuszak, G.} \& \bibinfo{author}{Abramowski, A.}
\newblock \bibinfo{title}{Algorithm and architecture design of the h. 265/hevc
  intra encoder}.
\newblock \emph{\bibinfo{journal}{IEEE Transactions on circuits and systems for
  video technology}} \textbf{\bibinfo{volume}{26}}, \bibinfo{pages}{210--222}
  (\bibinfo{year}{2015}).

\bibitem{coding2020iec}
\bibinfo{author}{Coding, V.~V.} \& \bibinfo{author}{Standard, I.}
\newblock \bibinfo{title}{Iec 23090-3}.
\newblock \emph{\bibinfo{journal}{ISO/IEC JTC}} \textbf{\bibinfo{volume}{1}}
  (\bibinfo{year}{2020}).

\bibitem{abu2016youtube}
\bibinfo{author}{Abu-El-Haija, S.} \emph{et~al.}
\newblock \bibinfo{title}{Youtube-8m: a large-scale video classification
  benchmark}.
\newblock \emph{\bibinfo{journal}{arXiv}}  (\bibinfo{year}{2016}).
\newblock \urlprefix\url{https://arxiv.org/abs/1609.08675}.

\bibitem{caba2015activitynet}
\bibinfo{author}{Caba~Heilbron, F.}, \bibinfo{author}{Escorcia, V.},
  \bibinfo{author}{Ghanem, B.}, \bibinfo{author}{Carlos~Niebles, J.} \&
  \bibinfo{author}{Ieee}.
\newblock \bibinfo{title}{Activitynet: A large-scale video benchmark for human
  activity understanding}.
\newblock In \emph{\bibinfo{booktitle}{IEEE Conference on Computer Vision and
  Pattern Recognition (CVPR)}}, \bibinfo{pages}{961--970}
  (\bibinfo{year}{2015}).

\bibitem{soomro2012ucf101}
\bibinfo{author}{Soomro, K.}, \bibinfo{author}{Zamir, A.~R.} \&
  \bibinfo{author}{Shah, M.}
\newblock \bibinfo{title}{Ucf101: A dataset of 101 human actions classes from
  videos in the wild}.
\newblock \emph{\bibinfo{journal}{arXiv}}  (\bibinfo{year}{2012}).
\newblock \urlprefix\url{https://arxiv.org/abs/1212.0402}.

\bibitem{karpathy2014large}
\bibinfo{author}{Karpathy, A.} \emph{et~al.}
\newblock \bibinfo{title}{Large-scale video classification with convolutional
  neural networks}.
\newblock In \emph{\bibinfo{booktitle}{27th IEEE Conference on Computer Vision
  and Pattern Recognition (CVPR)}}, \bibinfo{pages}{1725--1732}
  (\bibinfo{year}{2014}).

\bibitem{wiki}
\bibinfo{title}{List of most-subscribed youtube channels}.
\newblock
  \urlprefix\url{https://en.wikipedia.org/wiki/List_of_most-subscribed_YouTube_channels}.

\bibitem{wang2017time}
\bibinfo{author}{Wang, Z.}, \bibinfo{author}{Yan, W.}, \bibinfo{author}{Oates,
  T.} \& \bibinfo{author}{Ieee}.
\newblock \bibinfo{title}{Time series classification from scratch with deep
  neural networks: A strong baseline}.
\newblock In \emph{\bibinfo{booktitle}{International Joint Conference on Neural
  Networks (IJCNN)}}, \bibinfo{pages}{1578--1585} (\bibinfo{year}{2017}).

\bibitem{berndt1994using}
\bibinfo{author}{Berndt, D.~J.} \& \bibinfo{author}{Clifford, J.}
\newblock \bibinfo{title}{Using dynamic time warping to find patterns in time
  series}.
\newblock In \emph{\bibinfo{booktitle}{Proceedings of the 3rd international
  conference on knowledge discovery and data mining}},
  \bibinfo{pages}{359--370} (\bibinfo{year}{1994}).

\bibitem{gamboa2017deep}
\bibinfo{author}{Gamboa, J. C.~B.}
\newblock \bibinfo{title}{Deep learning for time-series analysis}.
\newblock \emph{\bibinfo{journal}{arXiv}}  (\bibinfo{year}{2017}).
\newblock \urlprefix\url{https://arxiv.org/abs/1701.01887}.

\bibitem{bagnall2017great}
\bibinfo{author}{Bagnall, A.}, \bibinfo{author}{Lines, J.},
  \bibinfo{author}{Bostrom, A.}, \bibinfo{author}{Large, J.} \&
  \bibinfo{author}{Keogh, E.}
\newblock \bibinfo{title}{The great time series classification bake off: a
  review and experimental evaluation of recent algorithmic advances}.
\newblock \emph{\bibinfo{journal}{Data Mining and Knowledge Discovery}}
  \textbf{\bibinfo{volume}{31}}, \bibinfo{pages}{606--660}
  (\bibinfo{year}{2017}).

\bibitem{ruiz2021great}
\bibinfo{author}{Ruiz, A.~P.}, \bibinfo{author}{Flynn, M.},
  \bibinfo{author}{Large, J.}, \bibinfo{author}{Middlehurst, M.} \&
  \bibinfo{author}{Bagnall, A.}
\newblock \bibinfo{title}{The great multivariate time series classification
  bake off: a review and experimental evaluation of recent algorithmic
  advances}.
\newblock \emph{\bibinfo{journal}{Data Mining and Knowledge Discovery}}
  \textbf{\bibinfo{volume}{35}}, \bibinfo{pages}{401--449}
  (\bibinfo{year}{2021}).

\bibitem{jeong2011weighted}
\bibinfo{author}{Jeong, Y.-S.}, \bibinfo{author}{Jeong, M.~K.} \&
  \bibinfo{author}{Omitaomu, O.~A.}
\newblock \bibinfo{title}{Weighted dynamic time warping for time series
  classification}.
\newblock \emph{\bibinfo{journal}{Pattern Recognition}}
  \textbf{\bibinfo{volume}{44}}, \bibinfo{pages}{2231--2240}
  (\bibinfo{year}{2011}).

\bibitem{stefan2012move}
\bibinfo{author}{Stefan, A.}, \bibinfo{author}{Athitsos, V.} \&
  \bibinfo{author}{Das, G.}
\newblock \bibinfo{title}{The move-split-merge metric for time series}.
\newblock \emph{\bibinfo{journal}{IEEE Transactions on Knowledge and Data
  Engineering}} \textbf{\bibinfo{volume}{25}}, \bibinfo{pages}{1425--1438}
  (\bibinfo{year}{2013}).

\bibitem{batista2014cid}
\bibinfo{author}{Batista, G. E. A. P.~A.}, \bibinfo{author}{Keogh, E.~J.},
  \bibinfo{author}{Tataw, O.~M.} \& \bibinfo{author}{de~Souza, V. M.~A.}
\newblock \bibinfo{title}{Cid: an efficient complexity-invariant distance for
  time series}.
\newblock \emph{\bibinfo{journal}{Data Mining and Knowledge Discovery}}
  \textbf{\bibinfo{volume}{28}}, \bibinfo{pages}{634--669}
  (\bibinfo{year}{2014}).

\bibitem{gorecki2013using}
\bibinfo{author}{Gorecki, T.} \& \bibinfo{author}{Luczak, M.}
\newblock \bibinfo{title}{Using derivatives in time series classification}.
\newblock \emph{\bibinfo{journal}{Data Mining and Knowledge Discovery}}
  \textbf{\bibinfo{volume}{26}}, \bibinfo{pages}{310--331}
  (\bibinfo{year}{2013}).

\bibitem{gorecki2014no}
\bibinfo{author}{Gorecki, T.} \& \bibinfo{author}{Luczak, M.}
\newblock \bibinfo{title}{Non-isometric transforms in time series
  classification using dtw}.
\newblock \emph{\bibinfo{journal}{Knowledge-Based Systems}}
  \textbf{\bibinfo{volume}{61}}, \bibinfo{pages}{98--108}
  (\bibinfo{year}{2014}).

\bibitem{lines2015time}
\bibinfo{author}{Lines, J.} \& \bibinfo{author}{Bagnall, A.}
\newblock \bibinfo{title}{Time series classification with ensembles of elastic
  distance measures}.
\newblock \emph{\bibinfo{journal}{Data Mining and Knowledge Discovery}}
  \textbf{\bibinfo{volume}{29}}, \bibinfo{pages}{565--592}
  (\bibinfo{year}{2015}).

\bibitem{rodriguez2004support}
\bibinfo{author}{Rodriguez, J.~J.}, \bibinfo{author}{Alonso, C.~J.} \&
  \bibinfo{author}{Maestro, J.~A.}
\newblock \bibinfo{title}{Support vector machines of interval-based features
  for time series classification}.
\newblock \emph{\bibinfo{journal}{Knowledge-Based Systems}}
  \textbf{\bibinfo{volume}{18}}, \bibinfo{pages}{171--178}
  (\bibinfo{year}{2005}).

\bibitem{ismail2019deep}
\bibinfo{author}{Fawaz, H.~I.}, \bibinfo{author}{Forestier, G.},
  \bibinfo{author}{Weber, J.}, \bibinfo{author}{Idoumghar, L.} \&
  \bibinfo{author}{Muller, P.-A.}
\newblock \bibinfo{title}{Deep learning for time series classification: a
  review}.
\newblock \emph{\bibinfo{journal}{Data Mining and Knowledge Discovery}}
  \textbf{\bibinfo{volume}{33}}, \bibinfo{pages}{917--963}
  (\bibinfo{year}{2019}).

\bibitem{zhao2017convolutional}
\bibinfo{author}{Zhao, B.}, \bibinfo{author}{Lu, H.}, \bibinfo{author}{Chen,
  S.}, \bibinfo{author}{Liu, J.} \& \bibinfo{author}{Wu, D.}
\newblock \bibinfo{title}{Convolutional neural networks for time series
  classification}.
\newblock \emph{\bibinfo{journal}{Journal of Systems Engineering and
  Electronics}} \textbf{\bibinfo{volume}{28}}, \bibinfo{pages}{162--169}
  (\bibinfo{year}{2017}).

\bibitem{serra2018towards}
\bibinfo{author}{Serr{\`a}, J.}, \bibinfo{author}{Pascual, S.} \&
  \bibinfo{author}{Karatzoglou, A.}
\newblock \bibinfo{title}{Towards a universal neural network encoder for time
  series}.
\newblock In \emph{\bibinfo{booktitle}{Artificial Intelligence Research and
  Development}}, \bibinfo{pages}{120--129} (\bibinfo{publisher}{IOS Press},
  \bibinfo{year}{2018}).

\bibitem{cui2016multi}
\bibinfo{author}{Cui, Z.}, \bibinfo{author}{Chen, W.} \& \bibinfo{author}{Chen,
  Y.}
\newblock \bibinfo{title}{Multi-scale convolutional neural networks for time
  series classification}.
\newblock \emph{\bibinfo{journal}{arXiv}}  (\bibinfo{year}{2016}).
\newblock \urlprefix\url{https://arxiv.org/abs/1603.06995}.

\bibitem{zheng2014time}
\bibinfo{author}{Zheng, Y.}, \bibinfo{author}{Liu, Q.}, \bibinfo{author}{Chen,
  E.}, \bibinfo{author}{Ge, Y.} \& \bibinfo{author}{Zhao, J.~L.}
\newblock \bibinfo{title}{Time series classification using multi-channels deep
  convolutional neural networks}.
\newblock In \emph{\bibinfo{booktitle}{International conference on web-age
  information management}}, \bibinfo{pages}{298--310}
  (\bibinfo{organization}{Springer}, \bibinfo{year}{2014}).

\bibitem{le2016data}
\bibinfo{author}{Le~Guennec, A.}, \bibinfo{author}{Malinowski, S.} \&
  \bibinfo{author}{Tavenard, R.}
\newblock \bibinfo{title}{Data augmentation for time series classification
  using convolutional neural networks}.
\newblock In \emph{\bibinfo{booktitle}{ECML/PKDD workshop on advanced analytics
  and learning on temporal data}} (\bibinfo{year}{2016}).

\bibitem{tanisaro2016time}
\bibinfo{author}{Tanisaro, P.} \& \bibinfo{author}{Heidemann, G.}
\newblock \bibinfo{title}{Time series classification using time warping
  invariant echo state networks}.
\newblock In \emph{\bibinfo{booktitle}{15th IEEE International Conference on
  Machine Learning and Applications (ICMLA)}}, \bibinfo{pages}{831--836}
  (\bibinfo{organization}{IEEE}, \bibinfo{year}{2016}).

\bibitem{c.dataset1}
\bibinfo{author}{Anthony, B.} \emph{et~al.}
\newblock \urlprefix\url{http://timeseriesclassification.com/}.

\bibitem{c.dataset2}
\bibinfo{author}{Mustafa, B.}
\newblock \urlprefix\url{http://www.mustafabaydogan.com/}.

\bibitem{he2016deep}
\bibinfo{author}{Kaiming, H.}, \bibinfo{author}{Xiangyu, Z.},
  \bibinfo{author}{Shaoqing, R.} \& \bibinfo{author}{Jian, S.}
\newblock \bibinfo{title}{Deep residual learning for image recognition}.
\newblock \emph{\bibinfo{journal}{2016 IEEE Conference on Computer Vision and
  Pattern Recognition (CVPR)}} \bibinfo{pages}{770--778}
  (\bibinfo{year}{2016}).

\end{thebibliography}


\section{Acknowledgments}

The work was supported by Shenzhen Boyan Tech. Ltd. The authors thank Yuchen Deng and Fengpu Pan for their valuable contributions in data collection and reviewing the paper.

\section{Author contributions}

Yuxing Han and Jiangtao Wen conceived 
the original ideas and the key algorithms. Yunan Ding collected
and annotated the data sets, and designed and conducted 
the experiments. Chen Ye Gan contributed to the final 
version of the manuscript.

\section{Competing interests}

The authors declare no competing interests.

\end{document}